\def\bm{{\mathbf m}}

\def\texitem#1{\par\smallskip\noindent\hangindent 25pt
               \hbox to 25pt {\hss #1 ~}\ignorespaces}

\documentclass{article}

\usepackage{todonotes}

\usepackage[preprint, nonatbib]{neurips_2022}

\usepackage[utf8]{inputenc} 
\usepackage[T1]{fontenc}    
\usepackage{hyperref}       
\usepackage{url}            
\usepackage{booktabs}       
\usepackage{amsfonts}       
\usepackage{nicefrac}       
\usepackage{microtype}      
\usepackage{xcolor}         
\usepackage{subcaption}
\usepackage{caption}
\usepackage{graphicx}
\usepackage{amsmath,amsfonts,amsthm,bm}
\usepackage[ruled,vlined]{algorithm2e}

\usepackage[bibencoding=utf8,sorting=none,maxnames=50,firstinits=false]{biblatex}
\newtheorem{mydef}{Definition}
\newtheorem{Theorem}{Theorem}

\newtheorem{lemma}{Lemma}

\theoremstyle{definition}
\newtheorem{remark}{Remark}

\title{Unsupervised Reward Shaping for a Robotic Sequential Picking Task from Visual Observations in a Logistics Scenario}

\author{%
  Vittorio Giammarino\thanks{Part of this work was done while Giammarino was an intern in the AI team at Pickle Robot Co., 1280 Cambridge St., Cambridge, Massachusetts 02139.} \\
  Division of Systems Engineering\\
  Boston University\\
  Boston, MA 02446, USA \\
  \texttt{vgiammar@bu.edu} \\
  \And
  Andrew J Meyer \\
  AI Team \\
  Pickle Robot Co. \\
  Cambridge, MA 02139, USA\\
  \texttt{aj@picklerobot.com}
  \And
  Kai Biegun \\
  AI Team \\
  Pickle Robot Co. \\
  Cambridge, MA 02139, USA\\
  \texttt{kai@picklerobot.com}
}

\begin{document}

\maketitle

\begin{abstract}
We focus on an unloading problem, typical of the logistics sector, modeled as a sequential pick-and-place task. In this type of task, modern machine learning techniques have shown to work better than classic systems since they are more adaptable to stochasticity and better able to cope with large uncertainties. More specifically, supervised and imitation learning have achieved outstanding results in this regard, with the shortcoming of requiring some form of supervision which is not always obtainable for all settings. On the other hand, reinforcement learning (RL) requires much milder form of supervision but still remains impracticable due to its inefficiency. In this paper, we propose and theoretically motivate a novel Unsupervised Reward Shaping algorithm from expert's observations which relaxes the level of supervision required by the agent and works on improving RL performance in our task. 
\end{abstract}

\section{Introduction}
Fuelled by the Covid-19 pandemic, logistics has endured great stress in recent years and accelerated the adoption of new technologies. This has led to a surge in demand for intelligent robots able to operate in unstructured environment. 
In this paper we focus on a sequential picking task for logistics setting (see Fig.~\ref{fig:task}). A robot manipulator needs to unload a stack of parcels having as unique source of input a RGB-D (Red, Green, Blue, Depth) image of the scene. 

Classic systems address this type of pick-and-place problems by integrating objects' pose estimation \cite{yoon2003real, zhu2014single} with scripted planning and motion control \cite{frazzoli2005maneuver}. While these systems are robust in structured and deterministic setting such as manufacturing, they have shown to be unsuitable for the unstructured, noisy and stochastic setting typical of logistics.
This has renewed the interest in end-to-end models mapping pixels to action which can be trained from data and achieve better performance under uncertainties.
In this regard, Reinforcement Learning (RL) \cite{sutton2018reinforcement} turned out capable of training end-to-end models accomplishing good results in pick and place manipulation tasks \cite{levine2016end, kalashnikov2018scalable}. However, classic RL still requires copious amount of data to converge which makes it impracticable for real world scenarios. 
In order to improve sample efficiency \cite{florence2019self}, recent works have integrated RL with object-centric assumptions such as keypoints \cite{manuelli2019kpam, kulkarni2019unsupervised}, embeddings \cite{jund2018optimization} or dense descriptors \cite{florence2018dense}. These representations are usually learned via supervised learning (SL) \cite{liu2011supervised} which often imposes tedious and expensive data collection burdens. Another line of research has recently focused on imitation learning (IL) or learning from demonstrations (LfD) \cite{zeng2020transporter, huang2022equivariant}. Despite the remarkable results and the need for a weaker form of supervision, IL and LfD remain in the SL spectrum and both observations and action/labels need to be provided to the learner.

In this work, we do another step towards successfully and efficiently learning picking strategies in a less supervised manner. We propose an algorithm based on Generative Adversarial Imitation Learning (GAIL) for unsupervised reward shaping in Deep RL which only requires expert observations. Note that the distinction between expert observations and demonstrations will be better formalized later in the paper; in simple terms we refer to demonstrations as state-action pairs; whereas, expert observations are only the states visited by the expert's policy.
To the best of our knowledge, this is the first work which solves a sequential picking task from visual observations without direct supervision. We compare our algorithm, Unsupervised Reward Shaping with Adversarial Imitation Learning from Observations (URSfO) with its RL-only counterpart and show improvement in asymptotic performance.   

\paragraph{Outline:} The remainder of the paper is organized as follows. The Preliminaries section presents a brief formal introduction to the main algorithmic techniques used throughout the paper, i.e. RL, IL and Inverse RL. The Environment section provides details about the sequential picking task and the simulator used to recreate the task. The Methods section describes the tested algorithms and provides theoretical motivations for the URSfO formulation. The Implementations and Results section deals with implementation details and reports the obtained experimental results. Finally, in the Conclusions section we discuss our results and provide interesting directions for future work.       

\paragraph{Notation:} Unless otherwise indicated, we use uppercase letters (e.g., $S_t$) for random variables, lowercase letters (e.g., $s_t$) for values of random variables, script letters (e.g., $\mathcal{S}$) for sets, and bold lowercase letters (e.g., $\bm{\theta}$) for vectors. Let $[t_1 : t_2]$ be the set of integers $t$ such that $t_1 \leq t \leq t_2$; we write $S_t$ such that $t_1 \leq t \leq t_2$ as $S_{t_1 : t_2}$. Finally, $\mathbb{E}[\cdot]$ represents expectation, $\mathbb{P}(\cdot)$ probability and $\mathbb{D}(\cdot, \cdot)$ divergence between distributions. 

\section{Preliminaries}

\paragraph*{Reinforcement Learning:}We consider an infinite-horizon discounted Markov Decision Process (MDP) defined by the tuple $(\mathcal{S}, \mathcal{A}, P, r, d_0, \gamma)$ where $\mathcal{S}$ is the set of states and $\mathcal{A}$ is the set of actions. $P:\mathcal{S}\times \mathcal{A} \rightarrow \Delta_{\mathcal{S}}$ is the transition probability function and $\Delta_{\mathcal{S}}$ denotes the space of probability distributions over $\mathcal{S}$. The function $r:\mathcal{S}\times \mathcal{A} \rightarrow \mathbb{R}$ maps state-action pairs to scalar rewards, $d_0\in\Delta_{\mathcal{S}}$ is the initial state distribution and $\gamma \in [0,1)$ the discount factor. We model the decision agent as a stationary policy $\pi:\mathcal{S}\rightarrow\Delta_{\mathcal{A}}$, where $\pi(a|s)$ is the probability of taking action $a$ in state $s$. The goal is to choose a policy that maximizes the expected total discounted reward $J(\pi)=\mathbb{E}_{\tau}[\sum_{t=0}^{\infty}\gamma^t r(s_t,a_t)]$, where $\tau = (s_0,a_0,s_1,a_1,\dots)$ is a trajectory sampled according to $s_0 \sim d_0$, $a_t\sim\pi(\cdot|s_t)$ and $s_{t+1}\sim P(\cdot|s_t,a_t)$. To streamline the notation we write $\tau \sim \pi$. A policy $\pi$ induces a normalized discounted state visitation distribution $d_{\pi}$, where $d_{\pi}(s) = (1-\gamma)\sum_{t=0}^{\infty}\gamma^t\mathbb{P}(s_t=s | d_0,\pi,P)$. We define the corresponding normalized discounted state-action visitation distribution as $\mu_{\pi}(s,a)= d_{\pi}(s)\pi(a|s)$ and the normalized discounted state-transition visitation distribution as $\rho_{\pi}(s,s')=d_{\pi}(s)\int_{\mathcal{A}} P(s'|s,\Bar{a})\pi(\Bar{a}|s) d\Bar{a}$. We denote the state value function of $\pi$ as $V^{\pi}(s) = \mathbb{E}_{\tau \sim \pi}[\sum_{t=0}^{\infty}\gamma^t r(s_t,a_t)|S_0=s]$, the state-action value function $Q^{\pi}(s,a) = \mathbb{E}_{\tau \sim \pi}[\sum_{t=0}^{\infty}\gamma^t r(s_t,a_t)|S_0=s, A_0=a]$ and the advantage function as $A^{\pi}(s,a) = Q^{\pi}(s,a) - V^{\pi}(s)$. Finally, when a function is parameterized with parameters $\bm{\theta} \in \varTheta \subset \mathbb{R}^k$ we write $\pi_{\bm{\theta}}$.

\paragraph{Imitation Learning:}
Given a task and an expert, IL infers the underlying expert's policy distribution via a set of expert demonstrations (state-action pairs). Assuming the expert's policy is parameterized by parameters $\bm{\theta}^*$, we refer to the process of estimating $\bm{\theta}^*$ through a finite sequence of expert demonstrations $\tau_E = (s_{0:T},a_{0:T})$ with $2 \leq T < \infty$ as IL. One way to formulate this problem is through maximum likelihood estimation:
\begin{equation}
    \max_{\bm{\theta}}  \mathcal{L}(\bm{\theta}),
    \label{eq:IL_max_likelihood}
\end{equation}
where $\mathcal{L}(\bm{\theta})$ denotes the log-likelihood and is equivalent to the logarithm of the joint probability of generating the expert demonstrations $\tau_E = \{s_0,a_0,s_1,a_1,\dots,s_T,a_T\}$, i.e.,
\begin{equation}
    \mathcal{L}(\bm{\theta}) = \log \mathbb{P}^{\bm{\theta}}_{d_0}(\tau_E).
    \label{eq:log_likelihood}
\end{equation}
$\mathbb{P}^{\bm{\theta}}_{d_0}(\tau_E)$ in \eqref{eq:log_likelihood} is defined as 
\begin{align}
    \begin{split}
        &\mathbb{P}^{\bm{\theta}}_{d_0}(\tau) = 
        d_0(s_0)\bigg[\prod_{t=0}^{T} \pi_{\bm{\theta}}(a_t|s_t) \bigg]\bigg[\prod_{t=0}^{T-1}P(s_{t+1}|s_t,a_t)\bigg].
    \end{split}
    \label{eq:Preliminary_joint_distribution}
\end{align}
Computing the logarithm of \eqref{eq:Preliminary_joint_distribution} and neglecting the elements not parameterized by $\bm{\theta}$ we obtain the following maximization problem
\begin{equation}
    \max_{\bm{\theta}}  \sum_{t=0}^{T}\log \big(\pi_{\bm{\theta}}(a_t|s_t) \big).
    \label{eq:IL_max_prob}
\end{equation}

\paragraph{Inverse Reinforcement Learning:} 
Given a task and an expert, inverse RL learns a reward function $r_{\bm{\chi}}$ via a finite set of expert demonstrations $\tau_E = (s_{0:T},a_{0:T})$, which allows us to estimate the expert's policy distribution over state-action pairs.\\
One way to formulate this problem is by alternating a classification step with a RL step as in \cite{abbeel2004apprenticeship}. In this regard, GAIL \cite{ho2016generative} accomplishes important results formulating the IRL problem as a $\min$-$\max$ game between a discriminator network $D_{\bm{\chi}}$ and the learner policy $\pi_{\bm{\theta}}$ as similarly done in Generative Adversarial Netowrks (GANs) \cite{goodfellow2014generative}. \\ 
More specifically, assume we have a set of expert demonstrations $\tau_E$, a set of trajectories $\tau$ generated by $\pi_{\bm{\theta}}$ and a discriminator network as $D_{\bm{\chi}}: \mathcal{S}\times\mathcal{A} \to (0,1)$. GAIL alternatively optimizes the following two objectives:
\begin{align}
    \max_{\bm{\chi}} \ \ \ &\mathbb{E}_{\tau \sim \pi_{\bm{\theta}}}[\log(D_{\bm{\chi}}(s,a))] + \mathbb{E}_{\tau_E}[\log(1 - D_{\bm{\chi}}(s,a))], \label{eq:IRL_disc}\\
    \min_{\bm{\theta}}\ \ \ &\mathbb{E}_{\tau \sim \pi_{\bm{\theta}}}[\log(D_{\bm{\chi}}(s,a))] \label{eq:IRL_policy}.
\end{align}
It has been shown that alternatively optimizing \eqref{eq:IRL_disc} and \eqref{eq:IRL_policy} is equivalent to minimizing the Jensen-Shannon divergence between $\mu_{\pi_{\bm{\theta}}}(s,a)$ and $\mu_{\pi_E}(s,a)$, i.e. we are inferring the expert's state-action visitation distribution. Note that, \eqref{eq:IRL_policy} is minimized using the policy gradient theorem in \cite{sutton1999policy}. 

\section{Environment}
In the following we provide a detailed description of our sequential picking task. The environment is based on PyBullet \cite{coumans2016pybullet} and all our code is freely accessible at our \href{https://github.com/VittorioGiammarino/Bootstrapping-RL-4-Sequential-Picking}{GitHub repository}\footnote{https://github.com/VittorioGiammarino/Bootstrapping-RL-4-Sequential-Picking}. 
\begin{figure}[ht]
    \centering
    \begin{subfigure}[t]{0.25\textwidth}
        \centering
        \includegraphics[width=3.5cm, height=3.5cm]{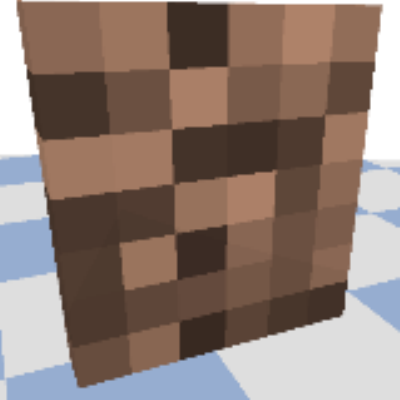}
        \caption{Visual Observation.}
        \label{fig:obs}
    \end{subfigure}
    ~
    \begin{subfigure}[t]{0.33\textwidth}
        \centering
        \includegraphics[width=6cm, height=3.5cm]{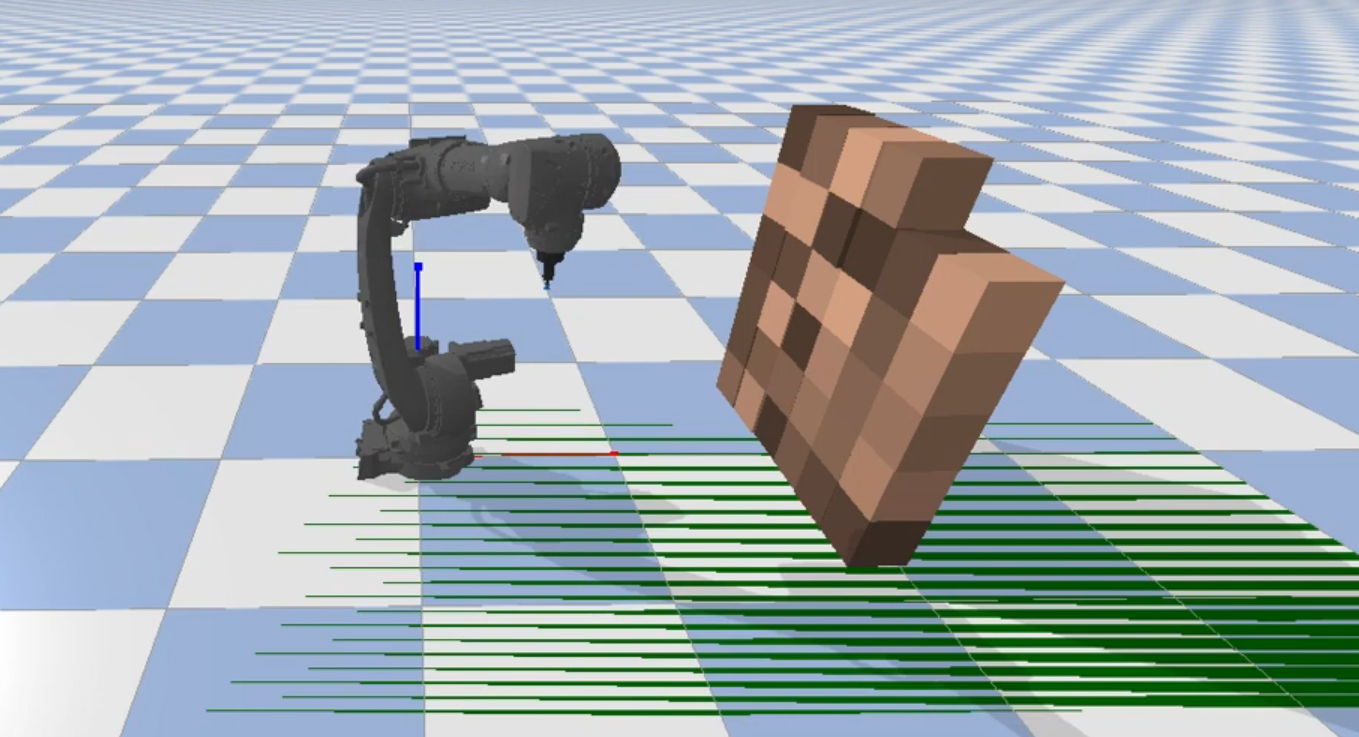}
        \caption{Environment.}
        \label{fig:task}
    \end{subfigure}
    ~
    \begin{subfigure}[t]{0.33\textwidth}
        \centering
        \includegraphics[width=3.6cm, height=3.6cm]{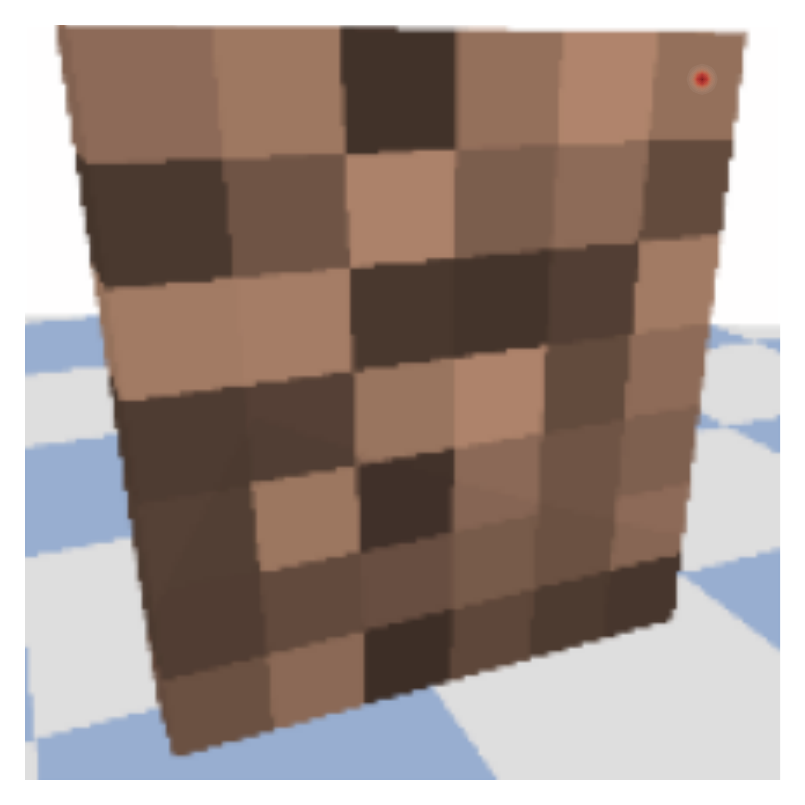}
        \caption{Visual Observation and Action.}
        \label{fig:obs+act}
    \end{subfigure}
    \caption{Overview of the Robotic setup. Video available \href{https://youtu.be/rcuXqNGOQRE}{here.}}
    \label{fig:task+obs}
\end{figure}

\begin{figure}
    \centering
    \includegraphics[width=\linewidth, height=4cm]{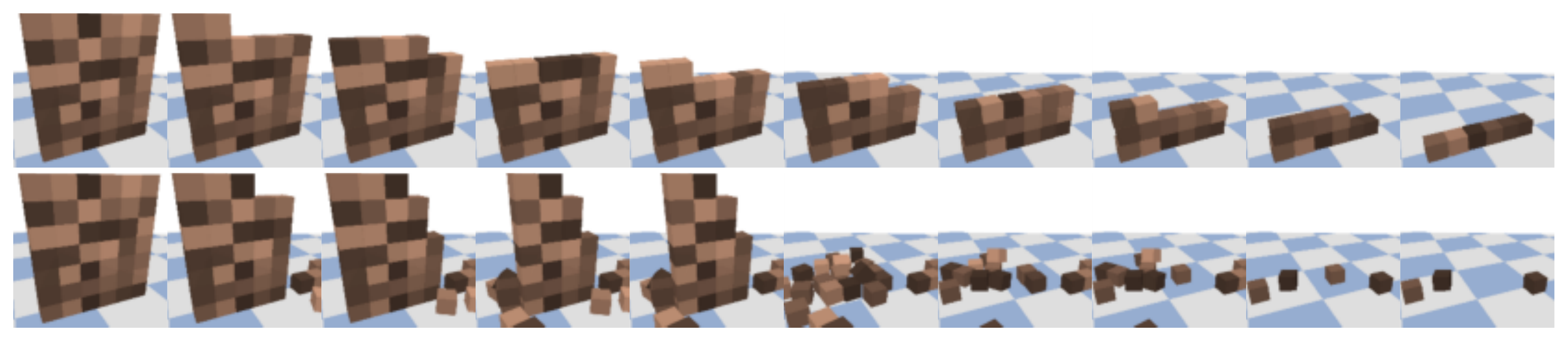}
    \caption{Evolution of the scene during two different episodes. In the upper episode the agent acts optimally and is able to unload almost all the parcels. On the other hand, the lower episode shows how a single wrong decision, halfway during the episode, irreversibly compromises the scene conditioning the agent's final outcome.}
    \label{fig:scene}
\end{figure}

\paragraph{Task:}We are interested in the sequential picking of stacks of parcels arranged as in Fig~\ref{fig:task+obs}. Our picking agent is an industrial KUKA KR70 robot manipulator equipped with a suction gripper. We consider the problem of learning sequential picking end-effector poses from visual observations $s_t$:
\begin{equation*}
    f(s_t) \to \mathcal{T}^{\text{pick}}_t, \ \ \  t \in (0,T),
\end{equation*}
where $\mathcal{T}^{\text{pick}}_t$ is the pose of the end-effector used to pick an object in SE(3) at time $t$ and $T$ is the maximum number of steps in a episode which corresponds to the total number of parcels in the scene. The placing pose $\mathcal{T}^{\text{place}}_t$ is predefined and corresponding to the same conveyor position for each box. The complexity of the task is given by the sequentiality of the decision process: the robot is required to correctly sequencing 42 picking actions in order to complete the task. See for instance Fig.~\ref{fig:scene} (lower episode), where a single wrong decision compromises the scene and consequently the agent's final outcome and success rate. Finally, note that the methods and algorithms presented in this work can be easily generalized beyond sequential picking to all the tasks consisting of sequences of two-pose motion primitives.

\paragraph{Perception and Control:}The robot's perception of the scene is based on a high resolution 720x1280 RGBD camera placed on the upper-left side of the robot frame. The camera captures the scene as in Fig.~\ref{fig:obs}; the RGB image is then cropped and resized to $160\times160$ pixels which represents the actual dimension of our visual observations space $\mathcal{S}$. \\
Further, we model our action space $\mathcal{A}$ as a discrete grid of $160\times160$ pixels $\{(u,v)\}$. At each timestep $t$, a pixel in the image is selected as in Fig.~\ref{fig:obs+act}, where the red spot in the top-right corresponds to the selected $\{(u,v)\}$. Having a depth image of the scene and knowing intrinsic and extrinsic parameters of the camera \cite{szeliski2010computer} we transform $\{(u,v)\}$ into robot frame xyz-coordinates. The obtained xyz-coordinates determine the picking or target position of the end-effector, while the target orientation is selected by hard-coded policy. More on this is deferred to the Primitives paragraph. Target position and orientation comprise the picking pose $\mathcal{T}^{\text{pick}}_t$. Given $\mathcal{T}^{\text{pick}}_t$, we solve an inverse kinematics optimization problem and recover the target configuration of the robot in joint space. In order to reach this target configuration the robot's joints are controlled in position using a PD controller.

\paragraph{Primitives:}The primitives used by the robot consists of $5$ main waypoints all characterized by a specific end-effector pose. The waypoints are: out-of-camera pose, pre-pick, $\mathcal{T}^{\text{pick}}_t$, post-pick, pre-place, and $\mathcal{T}^{\text{place}}_t$. The robot starts in a out-of-camera pose at $t=0$, it receives an image of the scene from the camera, selects $\mathcal{T}^{\text{pick}}_t$ and then moves following the aforementioned waypoints. After placing the parcel, the robot reaches again the out-of-camera pose, we set $t\leftarrow t+1$ and the transition starts again. The suction gripper is activated at $\mathcal{T}^{\text{pick}}_t$ provided a collision with an object is detected.\\
Furthermore, the task comes with two main modes: one forces the robot to attempt only side-picks which means picking only on the surfaces visible in \ref{fig:obs} and which are never entirely occluded in an ideal scenario. The second mode assumes the robot will attempt side picks for the parcels placed in the upper part of the pile and top-picks below $0.5m$. Where with top-picks we mean picking the top surface of a parcel. As a result, according to the z-coordinate of the selected picking position the primitive and the end-effector orientation in $\mathcal{T}^{\text{pick}}_t$ will change to facilitate top or side picks. Note that, the type of attempted pick will necessarily influence the optimal picking point yielding therefore different optimal policies for the two task modes.

\paragraph{Scene:}We define the scene as the pile of pickable parcels placed in front of the robot. Each parcel is a cube of $0.25m$ edge with randomized color within the brown spectrum. At the beginning of each episode the scene appears as in fig.~\ref{fig:obs}. The robot interacts with scene influencing the future parcels pose during the episode as illustrated in Fig.~\ref{fig:scene}. 

\section{Methods}
In the following, we formalize the end-to-end learning algorithms we tested on the sequential picking problem of the previous section. We present three main approaches which require different level of supervision, these are: Behavioral Cloning (BC) \cite{osa2018algorithmic}, Deep Q Learning (DQL) \cite{mnih2013playing} and our new algorithm called Unsupervised Reward Shaping using Adversarial Imitation Learning from observations (URSfO).

\paragraph{Behavioral Cloning:}
BC is a form of imitation learning from demonstrations which learns a direct mapping from states to actions. Given a set of demonstrations $\tau_E = \{s_0,a_0,s_1,a_1,\dots,s_T,a_T\}$ generated in a MDP, we define $\pi_{\bm{\theta}}: \mathcal{S} \to \Delta_{\mathcal{A}}$, and train $\pi_{\bm{\theta}}$ in a supervised fashion in order to clone the desired behavior. This approach treats the correlated state-action pairs generated in the MDP as identical and independent distributed (iid) leading to potential compounding of errors \cite{ross2010efficient}. However, being BC fundamentally a supervised learning algorithm, it turns quite effective when large amount of data are available and when associated with other techniques such as data augmentation and batch normalization.        
 
\paragraph{Deep Q-Learning:}
DQL is a model-free RL algorithm introduced in the seminal work in \cite{mnih2013playing}. Recall the definition of state-action value function $Q^{\pi}(s,a) = \mathbb{E}_{\tau \sim \pi}[\sum_{t=0}^{\infty}\gamma^t r(s_t,a_t)|S_0=s, A_0=a]$. We define a replay buffer $\mathcal{D}$ such that $(s,a,r,s')\sim\mathcal{D}$ and $Q_{\bm{\theta}}: \mathcal{S} \times \mathcal{A} \to \mathbb{R}$ and use the following updating rule based on the Bellman equation:
\begin{align}
    V_j &= \mathbb{E}_{(s_j,a_j,r_j,s'_j)\sim\mathcal{D}}[r_j + \gamma \max_{a'}Q_{\bm{\theta}}(s_j',a')], \label{eq:value} \\
    \bm{\theta}' &\in \arg\min_{\bm{\theta}} (V_j - Q_{\bm{\theta}}(s_j,a_j))^2. \label{eq:MSE}
\end{align}
In DQL with discrete $\mathcal{A}$, the value update in \eqref{eq:value} uses a greedy target. When the target is susceptible to error $\epsilon$, which is unavoidable when function approximation is used, we have $\mathbb{E}_{\epsilon}[\max_{a'}(Q(s',a')+\epsilon)] \geq \max_{a'}(Q(s',a')$ \cite{thrun1993issues}. This overestimation bias often destabilizes the training in practice. We address this issue by using Double DQL with delayed target network updates as originally proposed in \cite{van2016deep}.\\
Unlike BC, DQL learns the task requiring only a reward function as unique form of supervision. Note that, in our task it is trivial to specify a simple reward function. However, specifying an "informative" reward function which takes into account the uncertainties in the scene, the primitives design and the sequential nature of the decision process is a much harder task. 

\paragraph{Unsupervised reward shaping from expert's observations:}
As previously mentioned, BC assumes the existence of an expert which provides "ready-to-use" state-action pairs. On the other hand, DQL only requires designing a reward function. In the following we come up with an algorithm which merges the best of both worlds. We relax the assumption made for BC, requiring only the expert's observations $\tau_E^o=\{s_0, s_1, \dots, s_T\}$ rather than state-action pairs. Furthermore, we formalize an unsupervised reward shaping algorithm called URSfO which exploits $\tau_E^o$ in order to improve our reward function and make it more informative. 

Recall the expected total discounted reward as $J(\pi_{\bm{\theta}})=\mathbb{E}_{\tau}[\sum_{t=0}^{\infty}\gamma^t r(s_t,a_t)]$, the normalized discounted state visitation distribution as $d_{\pi_{\bm{\theta}}}(s) = (1-\gamma)\sum_{t=0}^{\infty}\gamma^t\mathbb{P}(s_t=s | d_0,\pi_{\bm{\theta}},P)$, the normalized discounted state-action visitation distribution as $\mu_{\pi_{\bm{\theta}}}(s,a)= d_{\pi_{\bm{\theta}}}(s)\pi_{\bm{\theta}}(a|s)$ and the state-transition visitation distribution as $\rho_{\pi_{\bm{\theta}}}(s,s')=d_{\pi_{\bm{\theta}}}(s)\int_{\mathcal{A}} P(s'|s,\Bar{a})\pi_{\bm{\theta}}(\Bar{a}|s) d\Bar{a}$. Given $\mathcal{A}$ as a discrete action space, we rewrite $\rho_{\pi_{\bm{\theta}}}(s, s') = d_{\pi_{\bm{\theta}}}(s)\sum_{\Bar{a}} P(s'|s,\Bar{a})\pi_{\bm{\theta}}(\Bar{a}|s)$. We further define the normalized discounted joint distribution as $\rho_{\pi_{\bm{\theta}}}(s, a, s') = \mu_{\pi_{\bm{\theta}}}(s,a)P(s'|s,a)$ and the density function of the inverse dynamics model $\rho_{\pi_{\bm{\theta}}}(a|s, s')$ as
\begin{equation}
    \rho_{\pi_{\bm{\theta}}}(a|s, s') = \frac{P(s'|s,a)\pi_{\bm{\theta}}(a|s)}{\sum_{\Bar{a}} P(s'|s,\Bar{a})\pi_{\bm{\theta}}(\Bar{a}|s)}.
\end{equation}
In line with prior work \cite{ghasemipour2020divergence}, we cast the learning from expert problem in a divergence minimization problem
\begin{equation*}
    \min_{\bm{\theta}} \ \ \ \mathbb{D}(\mu_{\pi_{\bm{\theta}}}(s,a), \mu_{\pi_E}(s,a)).
\end{equation*}
We can now formulate our main theoretical result which is instrumental for our algorithm.
\begin{Theorem}
If the agent and the expert share the same transition $P(s'|s,a)$ then the following inequality holds
\begin{equation*}
    \big|J(\pi_{E}) - J(\pi_{\bm{\theta}})\big| \leq \frac{\sqrt{2}R_{\max}}{1-\gamma}\Big(\sqrt{\mathbb{D}_{\chi^2}\big(\rho_{\pi_{\bm{\theta}}}(a|s, s')||\rho_{\pi_{E}}(a|s, s')\big) + \mathbb{D}_{\chi^2}\big(\rho_{\pi_{\bm{\theta}}}(s, s')||\rho_{\pi_{E}}(s, s')\big)}\Big)
\end{equation*}
where $\mathbb{D}_{\chi^2}$ is the Pearson $\chi^2$ divergence.

\begin{proof}
The proof follows from the definition of $J(\pi_{\bm{\theta}})$ and exploits a series of results provided in \cite{yang2019imitation} and basic inequalities between the Total Variation distance ($\mathbb{D}_{\text{TV}}$), the Kullback–Leibler divergence ($\mathbb{D}_{\text{KL}}$) and the Pearson $\chi^2$ divergence ($\mathbb{D}_{\chi^2}$). We provide the full derivation in appendix.
\end{proof}
\label{Theo:1}
\end{Theorem}

\begin{remark}
Theorem~\ref{Theo:1} suggests that when only expert's observations are available we cannot fully minimize the suboptimality of $\pi_{\bm{\theta}}$ by simply minimizing $\mathbb{D}_{\chi^2}\big(\rho_{\pi_{\bm{\theta}}}(s, s')||\rho_{\pi_E}(s, s')\big)$. Note that, when $P(s'|s,\Bar{a})$ is injective with respect to $\Bar{a}$, i.e. $P(s'|s,a) = 0 \ \ \ \forall a \in \mathcal{A}-\{\Bar{a}\}$, then $\mathbb{D}_{\chi^2}\big(\rho_{\pi_{\bm{\theta}}}(a|s, s')||\rho_{\pi_E}(a|s, s')\big)=0$ and therefore $\mathbb{D}_{\chi^2}\big(\rho_{\pi_{\bm{\theta}}}(s, s')||\rho_{\pi_E}(s, s')\big) = \mathbb{D}_{\chi^2}\big(\mu_{\pi_{\bm{\theta}}}(s, a)||\mu_{\pi_E}(s, a)\big)$ \cite{yang2019imitation}. This is not the case in our task since more than one picking point can potentially lead to the same transition.
\end{remark}

As a result, we propose an alternative algorithm which exploits $\tau_E^o=\{s_0, s_1, \dots, s_T\}$ only for reward shaping and not as the main objective. \\
We minimize $\mathbb{D}_{\chi^2}\big(\rho_{\pi_{\bm{\theta}}}(s, s')||\rho_{\pi_E}(s, s')\big)$ in Theorem~\ref{Theo:1} using the Least-Square Generative Adversarial Network (LSGAN) optimization scheme \cite{mao2017least} which has been proved to minimize $\mathbb{D}_{\chi^2}$ (see appendix~\ref{app:B}). More precisely, we define a discriminator $D_{\bm{\chi}}:\mathcal{S}\times \mathcal{S} \to \mathbb{R}$ and alternatively optimize the two following problems
\begin{align}
    \begin{split}
        \min_{\chi}& \ \ \ \frac{1}{2}\mathbb{E}_{\tau_E^o}\Big[\big(D_{\bm{\chi}}(s,s')-1\big)^2\Big] + \frac{1}{2}\mathbb{E}_{(s,s')\sim\rho_{\pi_{\bm{\theta}}}}\Big[\big(D_{\bm{\chi}}(s,s')+1\big)^2\Big] \\
        &+ \frac{\lambda}{2}\mathbb{E}_{\tau_E^o}\Big[\big|\big|\nabla_{\chi}D_{\bm{\chi}}(s,s')\big|\big|^2\Big],
        \label{eq:disc_LSGAN}
    \end{split} \\
    \min_{\bm{\theta}}& \ \ \ \frac{1}{2}\mathbb{E}_{(s,s')\sim\rho_{\pi_{\bm{\theta}}}}\Big[\big(D_{\bm{\chi}}(s,s')-1\big)^2\Big],
    \label{eq:policyLSGAN}
\end{align}
where a discriminator gradient penalty is added for improving training stability \cite{mescheder2018training}.

We can now focus on $\mathbb{D}_{\chi^2}\big(\rho_{\pi_{\bm{\theta}}}(a|s, s')||\rho_{\pi_E}(a|s, s')\big)$ in Theorem~\ref{Theo:1} which is intractable due to the unavailability of the expert's actions. We approximate this objective by using the following function:
\begin{equation}
    f(s,a) = \begin{cases}
    1-w \cdot \text{accuracy}, \ \ \ \text{if picking succeeds}, \\
    0, \ \ \ \text{otherwise},
    \end{cases}
    \label{eq:reward_func}
\end{equation}
where $w$ is an hyper-parameter and accuracy is defined as the euclidean distance between the picking point and center of mass of the parcel.
This function can be interpreted as an estimation of the reward function optimized by the expert and which is reasonable in our specific task of interest. Intuitively, \eqref{eq:reward_func} is likely maximized by an expert but it is hard to optimize in classic RL due to lack of feedback about the picking order. As a result, \eqref{eq:reward_func} will benefit from the information added by \eqref{eq:disc_LSGAN}-\eqref{eq:policyLSGAN}. 
Overall, the estimated reward function $r_{\text{tot}}(s,a)$ have the following form
\begin{equation}
    r_{\text{tot}}(s,a) = \lambda_1 f(s,a) + \lambda_2 \max[0,1-0.25(D_{\bm{\chi}}(s,s')-1)^2],
    \label{eq:reward}
\end{equation}
where we bound the second term to be in the range $[0,1]$ \cite{peng2021amp} and $\lambda_1$, $\lambda_2$ are tunable parameters. Note that, eq.~\ref{eq:reward} follows the additive reward shaping scheme as presented in \cite{ng1999policy}. URSfO is summarized in Algorithm~\ref{alg:URSfO}. 

\begin{algorithm}[H]
\label{alg:URSfO}
\caption{URSfO}
Input: $\tau_E^o$, $\pi_{\bm{\theta}}$, $D_{\bm{\chi}}$, $\lambda$, $\lambda_1$, $\lambda_2$, $w$ \\
\For{$k=1,K$}{
$\tau \sim \pi_{\bm{\theta}}$ \\
Optimize \eqref{eq:disc_LSGAN} \\
Use any RL algorithm to maximize $\mathbb{E}_{\tau}[\sum_{t=0}^{\infty}\gamma^t r_{\text{tot}}(s_t,a_t)]$ with $r_{\text{tot}}(s,a)$ in \eqref{eq:reward}
}
\Return $\pi_{\bm{\theta}}$
\end{algorithm}

\section{Implementations and Results}
Recall we define $\mathcal{S}$ as a $160\times160$ RGB pixel space and $\mathcal{A}$ as a discrete action space of dimension $160\times160$. All our algorithms will focus on learning the critic network $Q_{\bm{\theta}}:\mathcal{S}\times \mathcal{A} \to \mathbb{R}$  which is parameterized as $43$-layer encoder decoder residual network (ResNet) \cite{he2016deep} with $12$ residual blocks. We use three $2$-stride convolutions in the encoder and three bilinear-upsampling layers in the decoder as in \cite{zeng2020transporter}. Each layer, except the last, is interleaved with ReLU activation functions. Given $Q_{\bm{\theta}}$, $\pi_{\bm{\theta}}(s) \in \arg\max_{\tilde a} Q_{\bm{\theta}}(s,\tilde a)$.\\
For the discriminator $D_{\chi}$, we use a 4-layer convolutional network which is then followed by a feed-forward network with a single hidden layer. More details and other important hyperparameters are available in the appendix~\ref{app:hyper}. 

\begin{table}
\centering
\caption{Results for BC. We evaluate each policy trained for a different amount of training steps and with access to a different amount of expert's demonstrations. The evaluation occurs for $10$ randomized scenarios previously unseen by the agent and we report the percentage of successful picks over the entire evaluation, i.e. $420$ parcels. Side-picks-only and side-and-top-picks denote the task modes as described in the Environment section.}
\label{tab:table_offline}
\footnotesize
\begin{tabular}{c c c c c c c}\toprule
& \multicolumn{6}{c}{number of demonstrated episodes}\\
\cmidrule(lr){2-7}
& \multicolumn{3}{c}{side picks only} &  \multicolumn{3}{c}{side and top picks}\\
\cmidrule(lr){2-4} \cmidrule(lr){5-7}
training steps & 1 & 10 & 50 & 1 & 10 & 50\\
\bottomrule
100 & $100\%$ & $95\%$ & $100\%$ & $67\%$ & $95\%$ & $99\%$\\
1,000 & $98\%$ & $100\%$ & $100\%$ & $92\%$ & $90\%$ & $86\%$ \\
10,000 & $100\%$ & $97\%$ & $100\%$ & $94\%$ & $99\%$ & $100\%$\\
100,000& $100\%$ & $100\%$ & $99\%$ & $99\%$ & $96\%$ & $99\%$ \\
\bottomrule
\end{tabular}
\label{tab:BC}
\end{table}

\paragraph{Behavioral Cloning:} We start describing our results for BC. We use as training loss the cross-entropy between one-hot representations of the expert's actions and $Q_{\bm{\theta}}(s,\cdot)$ with a softmax as output layer. In this context we rewrite it as $f_{\bm{\theta}}(s):\mathcal{S}\to\Delta_{\mathcal{A}}$ and interpret it as a function that given a scene provides a distribution over picking points. We therefore minimize the following loss
\begin{equation*}
    \mathcal{L}(\bm{\theta}) = -\mathbb{E}_{\tau_E}[\log f_{\bm{\theta}}(s)].
\end{equation*}
We vary the experiments for a different amount of training steps and demonstrated episodes and then evaluate for $10$ episodes in unseen scenarios. The results, expressed in percentage of successful picks, are in Table~\ref{tab:BC} for both the task modes. Table~\ref{tab:BC} gives interesting insights about the complexity of the two modes: the side-picks-only is remarkably easier to solve in this setting than side-and-top-picks. This occurs due to a combination of factors such as camera pose, camera resolution, parcels dimension and overall ratio of top versus side picks in a demonstrated episode. More specifically, camera pose, resolution and parcels dimension influence the number of pixels for each box surface. Fewer pixels per surface greatly reduces the margin of picking error, thus increasing the probability of making mistakes. This is particularly relevant for top surfaces which, due to the camera pose, always appear smaller than side surfaces (Fig.~\ref{fig:scene}). At the same time, given we collect $\tau_E$ using a hard-coded policy based on the primitive described in the Modeling section, we have in the expert's data a higher number of demonstrated side picks rather than top picks. It follows that more training steps and/or demonstrations are needed when top picks are required. 

\begin{figure}
    \centering
    \includegraphics[width=0.7\linewidth]{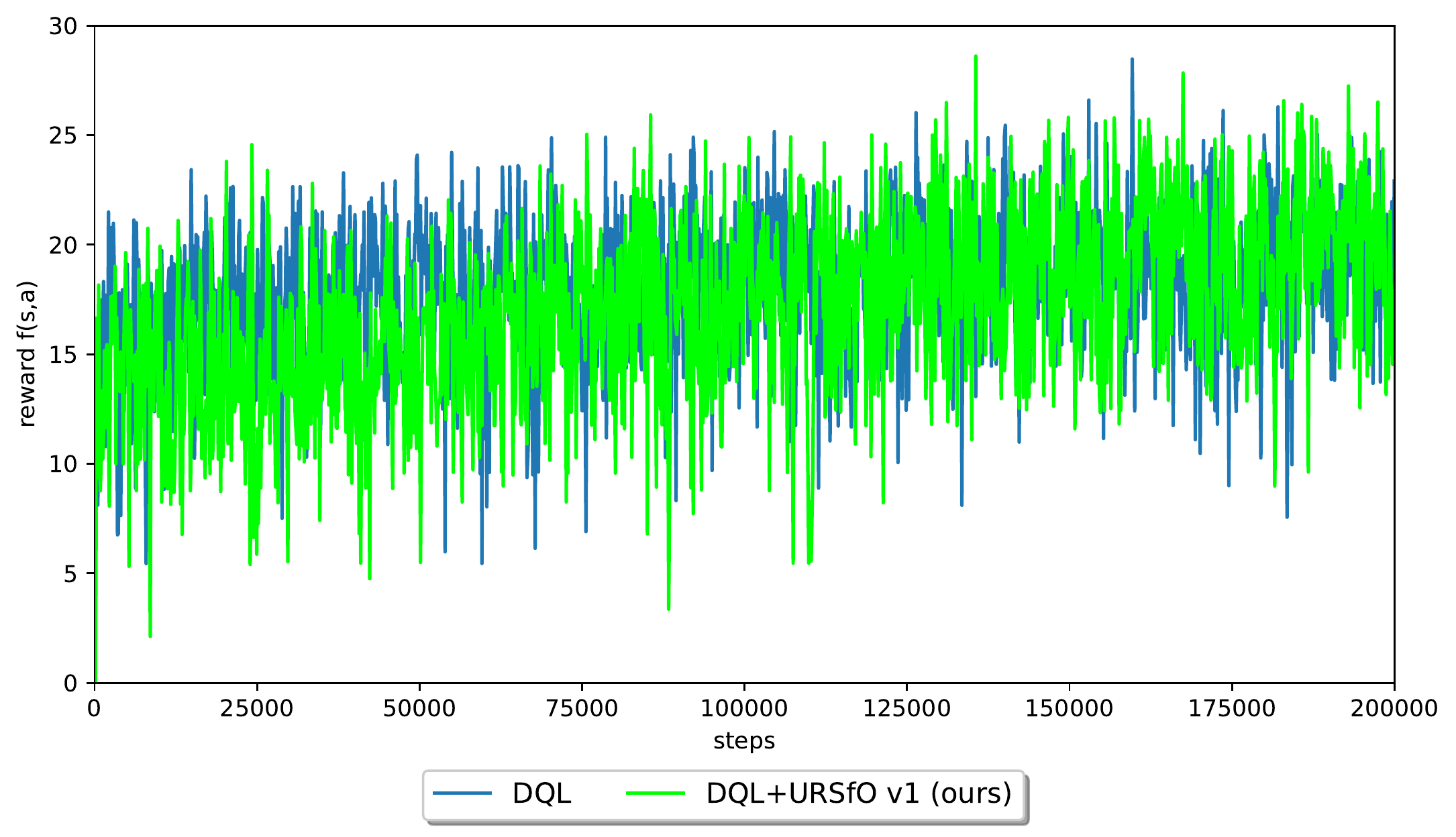}
    \caption{Results for the RL experiment. On the $x$-axis we show the training steps and on the $y$-axis the reward in \eqref{eq:reward_func} which represents an objective metrics to evaluate the agents' picking rate. We train both DQL and DQL+URSfO for 200k steps and evaluate the algorithms every $3$ full scenes, i.e. $126$ steps, for a single randomized scene (see fig.~\ref{fig:obs} for a scene example). DQL shows a slightly faster convergence in the first iterations which however quickly plateaus and oscillates. On the other hand, DQL+URSfO shows a more monotonic improvement over the training and better asymptotic performance. At the end of the full training, DQL+URSfO can successfully pick $34$ parcels versus the $25$ of DQL, a $21\%$ increase.}
    \label{fig:RL}
\end{figure}

\paragraph{Reinforcement Learning:} In the following we present our results for RL. We focus on the side-picks-only mode which has shown to be easily solvable assuming full expert supervision. We test both DQL and DQL+URSfO standardizing the two algorithms over the critic network, replay buffer and exploration strategy. The only source of difference is therefore the reward function: DQL will maximize the simple reward in \eqref{eq:reward_func}, whereas DQL+URSfO will maximize \eqref{eq:reward} with $\lambda_1=1$ and $\lambda_2=\delta$ where $\delta = 0$ at $t=0$ and $\delta = 2$ at $t=180$k with a linear increase over training steps (see Table~\ref{tab:Hyper_1} in appendix~\ref{app:hyper}). The idea behind this choice is to let both the discriminator network $D_{\bm{\chi}}$ and the policy network $\pi_{\bm{\theta}}$ to warm up independently during the first training instances and leverage $D_{\bm{\chi}}$ feedback only after the training has been stabilized. Fig.~\ref{fig:RL} shows the final results for this experiment. It is evident that the RL agents benefits throughout the whole training from maximizing \eqref{eq:reward} rather than \eqref{eq:reward_func}. Eventually, DQL+URSfO outperforms its DQL counterpart of roughly $21\%$ which means an additional $9$ parcels collected per scene.

\begin{figure}
    \centering
    \includegraphics[width=0.7\linewidth]{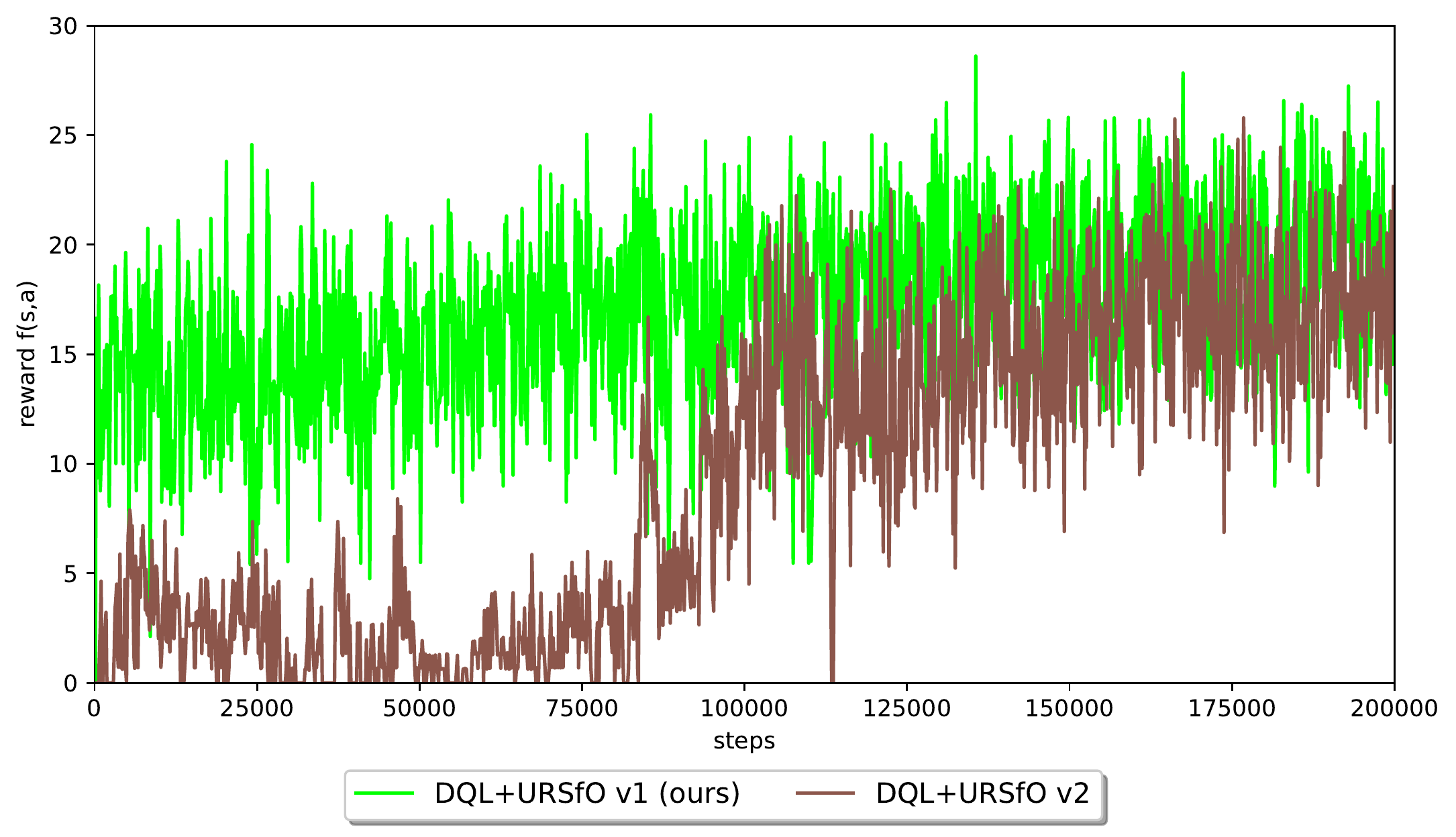}
    \caption{Results for the URSfO experiment with different $\lambda_2=\delta$. On the $x$-axis we show the training steps and on the $y$-axis the reward in \eqref{eq:reward_func} which represents an objective metrics to evaluate the agents' picking rate. We train DQL+URSfO with different $\lambda_2=\delta$ choices for 200k steps and evaluate the algorithms every $3$ full scenes, i.e. $126$ steps, for a single randomized scene (see fig.~\ref{fig:obs} for a scene example). DQL+URSfO~v1 sets $\delta = 0$ at $t=0$ and $\delta = 2$ at $t=180$k with a linear increase over steps. Whereas, DQL+URSfO~v2 sets $\delta = 2$ at $t=0$ and $\delta = 0$ at $t=180$k with linear decay over steps. DQL+URSfO~v2 suffers from the instability of adversarial learning during the first iterations and eventually shows lower performance than DQL+URSfO~v1.}
    \label{fig:URSfO_ablation}
\end{figure}

Finally, in order to validate our choice for $\lambda_2=\delta$, we test a decreasing version of $\delta$ where $\delta = 2$ at $t=0$ and $\delta = 0$ at $t=180$k with linear decay over training steps. The logic here is to exploit the expert knowledge during the first part of the training and then let slowly the agent on its own to avoid excessive and unnecessary bias towards the expert's behavior. Fig.~\ref{fig:URSfO_ablation} shows the comparison between this second version of our algorithm (DQL+URSfO~v2) and the original one (DQL+URSfO~v1). Interestingly, DQL+URSfO~v2 suffers from the instability of adversarial learning during the first training iterations and shows worse asymptotic performance than DQL+URSfO~v1.


\section{Conclusions}

In this paper, we model a robotic sequential picking task, typical of the logistics sector, and open source a Pybullet \href{https://github.com/VittorioGiammarino/Bootstrapping-RL-4-Sequential-Picking}{simulator} \cite{coumans2016pybullet} for prototyping and testing sequential decision algorithms on this task. Furthermore, we theoretically motivate a reward shaping algorithm based on adversarial IL which requires only visual observations rather than state-action pairs. In the experimental part of the paper, we test three different approaches on our task: IL, DQL and our formulated URSfO together with DQL (URSfO+DQL). As Table~\ref{tab:BC} shows, IL achieves remarkable results robustly hitting $99\%$ industrial standard of collected parcels. On the other hand, DQL and DQL+URSfO are both unable to reach the performance level of IL and remain in the range of $55\%$ to $95\%$ of parcels successfully picked per scene. Our URSfO provides a solid $21\%$ improvement over the DQL only approach (roughly $9$ parcels per scene) and hits peaks of $95\%$ picking success rate. Future work, will focus on developing and testing more principled tuning method for $\lambda_1$ and $\lambda_2$ in eq.~\ref{eq:reward} in an attempt to further improve both transient and asymptotic results. In addition to this, from an algorithmic point of view, we will search for alternatives to adversarial imitation learning and under the representation learning perspective we will explore alternatives to CNN based ResNets, such as attention based models \cite{dosovitskiy2020image}, for parameterizing $Q_{\bm{\theta}}$.


\printbibliography

\clearpage

\appendix

\section{Proof of Theorem~\ref{Theo:1}}

We start by introducing a series of definitions and lemmas which are instrumental for our proof. For sake of simplicity we consider both $\mathcal{S}$ and $\mathcal{A}$ to be discrete spaces.

\begin{mydef}[Total Variation Distance]
    Given $P$ and $Q$ discrete distribution over $\mathcal{X}$ the total variation between $P$ and $Q$ is
    \begin{equation}
        \mathbb{D}_{\text{TV}}(P,Q) = \frac{1}{2}\sum_{x\in \mathcal{X}}|P(x) - Q(x)|.
    \end{equation}
\end{mydef}

\begin{mydef}[KL Divergence]
    Given $P$ and $Q$ discrete distribution over $\mathcal{X}$ the KL divergence between $P$ and $Q$ is
    \begin{equation}
        \mathbb{D}_{\text{KL}}(P||Q) = \sum_{x\in \mathcal{X}}P(x)\log\frac{P(x)}{Q(x)}.
    \end{equation}
\end{mydef}

\begin{mydef}[Pearson $\chi^2$ Divergence]
    Given $P$ and $Q$ discrete distribution over $\mathcal{X}$ the Pearson $\chi^2$ Divergence between $P$ and $Q$ is
    \begin{equation}
        \mathbb{D}_{\chi^2}(P||Q) = \sum_{x\in \mathcal{X}}\frac{(P(x)-Q(x))^2}{Q(x)}.
    \end{equation}
\end{mydef}

\begin{lemma}[Relation between Distances]
\label{lemma_1}
\begin{equation}
    2(\mathbb{D}_{\text{TV}}(P,Q))^2 \leq \mathbb{D}_{\text{KL}}(P||Q) \leq \mathbb{D}_{\chi^2}(P||Q).
    \label{eq:lemma1}
\end{equation}
\begin{proof}
The first inequality follows from Pinsker's Inequality. The second can be proved as follows
\begin{align*}
    \mathbb{D}_{\chi^2}(P||Q) = \sum_{x\in \mathcal{X}}\frac{(P(x)-Q(x))^2}{Q(x)} = \sum_{x\in \mathcal{X}}\frac{P^2(x)}{Q(x)} - 1,
\end{align*}
where we simply expanded the expression and used $\sum_{x\in \mathcal{X}}P(x) = \sum_{x\in \mathcal{X}}Q(x) = 1$. Then, using $\log x < x - 1$ we get
\begin{align*}
    \mathbb{D}_{\text{KL}}(P||Q) = \sum_{x\in \mathcal{X}}P(x)\log\frac{P(x)}{Q(x)} \leq  \sum_{x\in \mathcal{X}}P(x)\bigg(\frac{P(x)}{Q(x)} -1 \bigg) = \mathbb{D}_{\chi^2}(P||Q).
\end{align*}
\end{proof}
\end{lemma}

\begin{lemma}[\cite{yang2019imitation}]
\label{lemma_2}
Under the assumption that the transition $P(s'|s,a)$ is stationary between the expert and the agent then the following equality holds
\begin{equation}
    \mathbb{D}_{\text{KL}}[\rho_{\pi_{\bm{\theta}}}(s, a, s')||\rho_{\pi_E}(s, a, s')] = \mathbb{D}_{\text{KL}}[\mu_{\pi_{\bm{\theta}}}(s, a)||\mu_{\pi_E}(s, a)].
\end{equation}

\begin{proof}
\begin{align*}
    \mathbb{D}_{\text{KL}}[\rho_{\pi_{\bm{\theta}}}(s, a, s')||\rho_{\pi_E}(s, a, s')] &= \mathbb{E}_{\rho_{\pi_{\bm{\theta}}}}\bigg[\log \frac{\rho_{\pi_{\bm{\theta}}}(s, a, s')}{\rho_{\pi_E}(s, a, s')}\bigg] \\
    &= \mathbb{E}_{\rho_{\pi_{\bm{\theta}}}}\bigg[\log \frac{\mu_{\pi_{\bm{\theta}}}(s,a)P(s'|s,a)}{\mu_{\pi_E}(s,a)P(s'|s,a)}\bigg] \\
    &= \mathbb{D}_{\text{KL}}[\mu_{\pi_{\bm{\theta}}}(s, a)||\mu_{\pi_E}(s, a)].
\end{align*}
\end{proof}
\end{lemma}

\begin{lemma}[\cite{yang2019imitation}]
Under the assumption that the transition $P(s'|s,a)$ is stationary between the expert and the agent then the following holds
\begin{equation}
    \mathbb{D}_{\text{KL}}[\mu_{\pi_{\bm{\theta}}}(s, a)||\mu_{\pi_E}(s, a)] = \mathbb{D}_{\text{KL}}[\rho_{\pi_{\bm{\theta}}}(a|s, s')||\rho_{\pi_E}(a| s, s')] + \mathbb{D}_{\text{KL}}[\rho_{\pi_{\bm{\theta}}}(s, s')||\rho_{\pi_E}(s, s')].
\end{equation}
\begin{proof}
\begin{align*}
     &\mathbb{D}_{\text{KL}}[\rho_{\pi_{\bm{\theta}}}(a|s, s')||\rho_{\pi_E}(a| s, s')] + \mathbb{D}_{\text{KL}}[\rho_{\pi_{\bm{\theta}}}(s, s')||\rho_{\pi_E}(s, s')] \\
     &=\sum_s \sum_a \sum_{s'} \rho_{\pi_{\bm{\theta}}}(s, a, s') \bigg(\log \frac{\rho_{\pi_{\bm{\theta}}}(a|s, s')}{\rho_{\pi_E}(a| s, s')} + \log \frac{\rho_{\pi_{\bm{\theta}}}(s, s')}{\rho_{\pi_E}(s, s')}\bigg) \\
     &= \mathbb{D}_{\text{KL}}[\rho_{\pi_{\bm{\theta}}}(s, a, s')||\rho_{\pi_E}(s, a, s')].
\end{align*}
The final result follows from lemma~\ref{lemma_2}.
\end{proof}
\end{lemma}

Now we can prove Theorem~\ref{Theo:1}:
\paragraph{Theorem~\ref{Theo:1}:}If the agent and the expert share the same transition $P(s'|s,a)$ then the following inequality holds
\begin{equation*}
    \big|J(\pi_{E}) - J(\pi_{\bm{\theta}})\big| \leq \frac{\sqrt{2}R_{\max}}{1-\gamma}\Big(\sqrt{\mathbb{D}_{\chi^2}\big(\rho_{\pi_{\bm{\theta}}}(a|s, s')||\rho_{\pi_E}(a|s, s')\big) + \mathbb{D}_{\chi^2}\big(\rho_{\pi_{\bm{\theta}}}(s, s')||\rho_{\pi_E}(s, s')\big)}\Big)
\end{equation*}
where $\mathbb{D}_{\chi^2}$ is the Pearson $\chi^2$ divergence.
\begin{proof}
\begin{align*}
    \big|J(\pi_{E}) - J(\pi_{\bm{\theta}})\big| &= \frac{1}{1-\gamma}\bigg|\sum_a\sum_s\mu_{\pi_E}(s, a)r(s,a) - \sum_a\sum_s\mu_{\pi_{\bm{\theta}}}(s, a)r(s,a)\bigg| \\ 
    &\leq \frac{R_{\max}}{1-\gamma} \sum_a\sum_s \Big| \mu_{\pi_E}(s, a) - \mu_{\pi_{\bm{\theta}}}(s, a)\Big| \\
    &= \frac{2R_{\max}}{1-\gamma} \mathbb{D}_{\text{TV}}(\mu_{\pi_E}(s,a),\mu_{\pi_{\bm{\theta}}}(s,a)).
\end{align*}
By Pinsker's inequality (lemma~\ref{lemma_1}) and lemma~\ref{lemma_2} it follows that 
\begin{align*}
    \big|J(\pi_{E}) - J(\pi_{\bm{\theta}})\big| &\leq \frac{\sqrt{2}R_{\max}}{1-\gamma} \sqrt{\mathbb{D}_{\text{KL}}[\mu_{\pi_E}(s,a)||\mu^{\pi_{\bm{\theta}}}(s,a)]} \\
    &\leq \frac{\sqrt{2}R_{\max}}{1-\gamma}\bigg(\sqrt{\mathbb{D}_{\text{KL}}[\rho_{\pi_{\bm{\theta}}}(a|s, s')||\rho_{\pi_E}(a| s, s')] + \mathbb{D}_{\text{KL}}[\rho_{\pi_{\bm{\theta}}}(s, s')||\rho_{\pi_E}(s, s')]}\bigg).
\end{align*}
The proof is concluded by applying again lemma~\ref{lemma_1}.
\end{proof}

\section{LSGAN and chi-squared divergence minimization}
\label{app:B}

Without loss of generality, we rewrite \eqref{eq:disc_LSGAN}-\eqref{eq:policyLSGAN} as
\begin{align*}
    \begin{split}
        \min_{\chi}& \ \ \ \frac{1}{2}\mathbb{E}_{\rho_{\pi_E}}\Big[\big(D_{\bm{\chi}}(s,s')-b\big)^2\Big] + \frac{1}{2}\mathbb{E}_{\rho_{\pi_{\bm{\theta}}}}\Big[\big(D_{\bm{\chi}}(s,s')-a\big)^2\Big],
    \end{split} \\
    \min_{\bm{\theta}}& \ \ \ \frac{1}{2}\mathbb{E}_{\rho_{\pi_E}}\Big[\big(D_{\bm{\chi}}(s,s')-c\big)^2\Big] + \frac{1}{2}\mathbb{E}_{\rho_{\pi_{\bm{\theta}}}}\Big[\big(D_{\bm{\chi}}(s,s')-c\big)^2\Big].
\end{align*}

The optimal $D_{\bm{\chi}}^*$ for fixed $\pi_{\bm{\theta}}$ is
\begin{equation*}
    D_{\bm{\chi}}^* = \frac{b\rho_{\pi_E}(s,s') + a\rho_{\pi_{\bm{\theta}}}(s,s')}{\rho_{\pi_E}(s,s') + \rho_{\pi_{\bm{\theta}}}(s,s')}.
\end{equation*}

We write
\begin{align*}
    2C(\pi_{\bm{\theta}}) =& \mathbb{E}_{\rho_{\pi_E}}\Big[\big(D_{\bm{\chi}}^*(s,s')-c\big)^2\Big] + \mathbb{E}_{\rho_{\pi_{\bm{\theta}}}}\Big[\big(D_{\bm{\chi}}^*(s,s')-c\big)^2\Big] \\
    =& \mathbb{E}_{\rho_{\pi_E}}\Bigg[\bigg(\frac{b\rho_{\pi_E}(s,s') + a\rho_{\pi_{\bm{\theta}}}(s,s')}{\rho_{\pi_E}(s,s') + \rho_{\pi_{\bm{\theta}}}(s,s')}-c\bigg)^2\Bigg] + \mathbb{E}_{\rho_{\pi_{\bm{\theta}}}}\Bigg[\bigg(\frac{b\rho_{\pi_E}(s,s') + a\rho_{\pi_{\bm{\theta}}}(s,s')}{\rho_{\pi_E}(s,s') + \rho_{\pi_{\bm{\theta}}}(s,s')}-c\bigg)^2\Bigg]\\
    =& \sum_s\sum_{s'}\rho_{\pi_E}(s,s')\bigg(\frac{(b-c)\rho_{\pi_E}(s,s') - (a-c)\rho_{\pi_{\bm{\theta}}}(s,s')}{\rho_{\pi_E}(s,s') + \rho_{\pi_{\bm{\theta}}}(s,s')}\bigg)^2 \\
    &+ \sum_s\sum_{s'}\rho_{\pi_{\bm{\theta}}}(s,s')\bigg(\frac{(b-c)\rho_{\pi_E}(s,s') - (a-c)\rho_{\pi_{\bm{\theta}}}(s,s')}{\rho^{\pi_E}(s,s') + \rho_{\pi_{\bm{\theta}}}(s,s')}\bigg)^2 \\
    =& \sum_s\sum_{s'}\frac{\Big((b-c)\rho_{\pi_E}(s,s') - (a-c)\rho_{\pi_{\bm{\theta}}}(s,s')\Big)^2}{\rho_{\pi_E}(s,s') + \rho_{\pi_{\bm{\theta}}}(s,s')} \\
    =& \sum_s\sum_{s'}\frac{\Big((b-c)(\rho_{\pi_E}(s,s')+\rho_{\pi_{\bm{\theta}}}(s,s')) - (b-a)\rho_{\pi_{\bm{\theta}}}(s,s')\Big)^2}{\rho_{\pi_E}(s,s') + \rho_{\pi_{\bm{\theta}}}(s,s')}.
\end{align*}
If we set $b-c=1$ and $b-a=2$, then
\begin{align*}
    2C(\pi_{\bm{\theta}}) =& \sum_s\sum_{s'}\frac{\Big(2\rho_{\pi_{\bm{\theta}}}(s,s') - (\rho_{\pi_E}(s,s')+\rho_{\pi_{\bm{\theta}}}(s,s'))\Big)^2}{\rho_{\pi_E}(s,s') + \rho_{\pi_{\bm{\theta}}}(s,s')}\\
    =& \mathbb{D}_{\chi^2}(2\rho_{\pi_{\bm{\theta}}}(s,s')||\rho_{\pi_E}(s,s') + \rho_{\pi_{\bm{\theta}}}(s,s')).
\end{align*}

\section{Hyperparameters}
\label{app:hyper}
\begin{table}[h!]
\centering
\caption{Hyperparameter values for experimental setup.}
\label{tab:Hyper_1}
\small
\begin{tabular}{c c c c}\toprule
\multicolumn{2}{l}{IL} & &\\
\cmidrule(lr){1-2}
\multicolumn{2}{l}{Optimization algorithm} & \multicolumn{2}{c}{Gradient ascent} \\
\multicolumn{2}{l}{Size minibatches} & \multicolumn{2}{c}{1} \\
\multicolumn{2}{l}{Optimizer} & \multicolumn{2}{c}{Adam}\\
\multicolumn{2}{l}{Learning rate} & \multicolumn{2}{c}{$1\times 10^{-4}$}\\
& & & \\
\multicolumn{2}{l}{DQL} & & \\
\cmidrule(lr){1-2}
\multicolumn{2}{l}{Number of seed steps} & \multicolumn{2}{c}{$100$} \\
\multicolumn{2}{l}{Number of training steps} & \multicolumn{2}{c}{$2\times 10^5$}\\
\multicolumn{2}{l}{Replay Buffer size} & \multicolumn{2}{c}{$2\times 10^5$}\\
\multicolumn{2}{l}{$w$} & \multicolumn{2}{c}{$2$} \\
\multicolumn{2}{l}{$\gamma$} & \multicolumn{2}{c}{0.99} \\
\multicolumn{2}{l}{Optimization algorithm} & \multicolumn{2}{c}{Gradient ascent} \\
\multicolumn{2}{l}{Size minibatches} & \multicolumn{2}{c}{16} \\
\multicolumn{2}{l}{Optimizer} & \multicolumn{2}{c}{Adam}\\
\multicolumn{2}{l}{Learning rate} & \multicolumn{2}{c}{$1\times 10^{-4}$}\\
\multicolumn{2}{l}{Update every} & \multicolumn{2}{c}{$2$ steps}\\
\multicolumn{2}{l}{Critic target $\tau$} & \multicolumn{2}{c}{$2\times 10^{-2}$}\\
& & & \\
\multicolumn{2}{l}{URSfO} & & \\
\cmidrule(lr){1-2}
\multicolumn{2}{l}{Discriminator feature dim} & \multicolumn{2}{c}{$100$}\\
\multicolumn{2}{l}{Discriminator hidden dim} & \multicolumn{2}{c}{$1024$}\\
\multicolumn{2}{l}{$\lambda_1$} & \multicolumn{2}{c}{$1$}\\
\multicolumn{2}{l}{$\lambda_2$} & \multicolumn{2}{c}{linear from $0$ to $2$ in $1.8\times 10^5$ steps}\\
\bottomrule
\end{tabular}
\end{table}

\end{document}